\pdfoutput=1

\documentclass[11pt]{article}
\usepackage{authblk}
\usepackage{graphicx}

\usepackage{acl}

\usepackage{times}
\usepackage{amsmath}
\usepackage{latexsym}
\usepackage{subfigure}
\usepackage{multirow}
\usepackage{relsize}
\usepackage[T1]{fontenc}

\usepackage[utf8]{inputenc}

\usepackage{microtype}
\usepackage{xspace}

\newcommand{\KPRS}{KPRS\xspace}

\definecolor{newgreen}{HTML}{196f3d}

\newcommand*\samethanks[1][\value{footnote}]{\footnotemark[#1]}

\title{Injecting Domain Knowledge in Language Models\\for Task-Oriented Dialogue Systems}

\author[1]{\thanks{~~~Work performed while at AWS AI Labs}  Denis Emelin}
\author[2]{Daniele Bonadiman}
\author[3,4]{\samethanks[1] Sawsan Alqahtani}
\author[2]{Yi Zhang}
\author[2]{Saab Mansour}

\affil[1]{University of Edinburgh \\ \texttt{denis.emelin@gmail.com}}
\affil[2]{AWS AI Labs \\ \texttt{\{dbonadim,yizhngn,saabm\}@amazon.com}}
\affil[3]{Princess Nourah Bint Abdulrahman \\ \texttt{saalqhtani@pnu.edu.sa}}
\affil[4]{National Center of AI \\ \texttt{sawalqahtani@nic.gov.sa}}

\begin{document}
\maketitle
\begin{abstract}
Pre-trained language models (PLM) have advanced the state-of-the-art across NLP applications, but lack domain-specific knowledge that does not naturally occur in pre-training data. Previous studies augmented PLMs with symbolic knowledge for different downstream NLP tasks. However, knowledge bases (KBs) utilized in these studies are usually large-scale and static, in contrast to small, domain-specific, and modifiable knowledge bases that are prominent in real-world task-oriented dialogue (TOD) systems. In this paper, we showcase the advantages of injecting domain-specific knowledge prior to fine-tuning on TOD tasks. To this end, we utilize light-weight adapters that can be easily integrated with PLMs and serve as a repository for facts learned from different KBs. To measure the efficacy of proposed knowledge injection methods, we introduce \textit{Knowledge Probing using Response Selection} (\KPRS) -- a probe designed specifically for TOD models. Experiments\footnote{\url{https://github.com/amazon-research/domain-knowledge-injection}} on \KPRS and the response generation task show improvements of knowledge injection with adapters over strong baselines.
\end{abstract}

\section{Introduction}
\label{introduction}

Pre-trained language models (PLMs), such as BERT \cite{devlin2018bert}, BART \cite{lewis-etal-2020-bart}, GPT \cite{brown2020language}, and XLNet \cite{yang2019xlnet}, have advanced the state-of-the-art of various natural language processing (NLP) technologies and demonstrated an exceptional ability to store and utilize linguistic, factual, and commonsense knowledge. Consequently, PLMs form the backbone of many recent NLP applications and have been successfully employed as modular components in the context of task-oriented dialogue (TOD), responsible for sub-tasks including dialogue state tracking and response generation \cite{hosseini2020simple,lee2021dialogue}. 

\begin{figure}[!t]
\centering
\includegraphics[width=0.9\columnwidth]{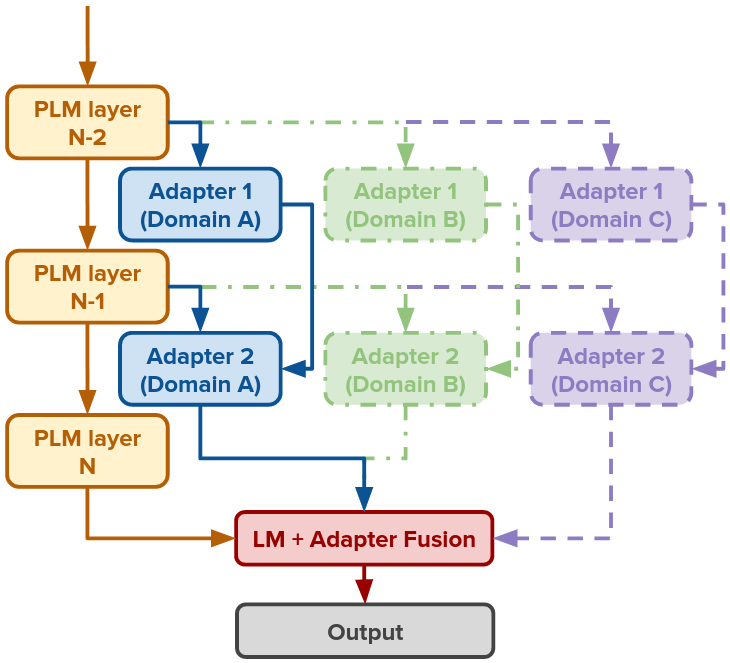}
\caption{A high-level representation of the KB-adapter architecture (decoder only, for clarity). Adapter states are fused with the hidden states of the PLM to produce a knowledge-informed predictive distribution.
Dashed elements are used only if multiple adapters are active.} 
\label{fig:kb_arch_fig}
\end{figure}

Since they are exposed to large quantities of general data during training, PLMs store a wide variety of diverse and general knowledge in their parameters \cite{petroni2019language} such as capitals of nations, biographical details of famous individuals, and other facts of varying granularity. Commercially deployed TOD systems, however, typically require access to more restricted, domain-specific categories of knowledge in order to produce informative and factually accurate responses to user queries.\footnote{The term \textit{domain} here refers to a specific application use-case (e.g. \texttt{expedia.com} (travel) and \texttt{opentable.com} (restaurant) represent different domains).} Such information may include addresses of particular local attractions, detailed restaurant menus, train routes, or ticket prices, and is unlikely to be found in the PLM's training data. Due to its specialized nature, this knowledge is often stored in external knowledge bases (KBs) that are accessed at run-time by TOD systems via external queries. 

This process introduces additional complexity into the dialogue model design and requires implementing KB queries and code wrappers as part of system setup, causing a substantial overhead especially for non-experts. Querying external KBs can also be disadvantageous when the KB is small, or is not changing in real time (as is the case with catalogs, restaurants' menus, etc). We identify the decoupling of domain-specific knowledge from the dialogue model as a shortcoming to be remedied and instead propose to inject this knowledge directly into the model's parameters. This eliminates the need for querying external KBs, streamlining the creation and deployment of TOD systems. 

Injecting domain-specific information into TOD systems that can guide and inform model behavior and may be subsequently updated and modified by the user is not a trivial task. Ideally, this should be accomplished in a manner that is efficient, architecture-agnostic, and compatible with off-the-shelf PLMs. In order to satisfy these requirements, we adopt \textbf{light-weight adapter networks as repositories of domain-specific knowledge} (\textbf{KB-adapters} for short). Such adapters can be trained to memorize KB facts\footnote{We use the term \textit{fact} to refer to individual KB entries.} and integrated into pretrained PLMs through the fusion of hidden representations, as illustrated in Figure \ref{fig:kb_arch_fig}.
Our work is in line with past studies that demonstrated the utility of adapters as stores of factual and linguistic knowledge outside of TOD \cite{wang2020k}. Importantly, injecting knowledge into TOD models through adapters is computationally less demanding than injecting domain-specific facts by fine-tuning entire dialogue models on synthetic data, as explored in \cite{madotto2020learning}, which facilitates efficient updating of the injected knowledge.

To quantify the success of the knowledge injection procedure, we develop the \textbf{K}nowledge \textbf{P}robing using \textbf{R}esponse \textbf{S}election (\KPRS) task and benchmark (see \S  \ref{knowledge_injection_benchmark}). \KPRS leverages contrastive dialogue response pairs to probe the extent of memorization of domain-specific facts by the evaluated dialogue model, whereby one response is consistent with the corresponding KB, while the other is not.

To our knowledge, both \KPRS and the use of adapters for domain-specific knowledge injection in TOD represent novel contributions of our work. We conduct experiments that evaluate PLMs equipped with domain-specific KB-adapters on the \KPRS benchmark as well as the more conventional response generation (RG) task, comparing them against strong baselines. 

Our contributions can be summarized as follows: 
\begin{itemize}
    \item We define and implement adapter-based methods for injecting highly specific and retrievable domain knowledge into TOD models
    \item We design and develop the \KPRS probing task that can be used to evaluate the effectiveness of knowledge injection for TOD systems
    \item We show that PLMs with KB-adapters are usually preferable to knowledge-unaware and sequentially-finetuned PLMs for TOD
\end{itemize}

\section{KB-Adapters for Domain-Specific Knowledge Injection}
\label{kbadapter}

We conceptualize KB adapters as repositories of domain-specific information that guide the PLMs' predictions to be consistent with KB contents. The proposed knowledge injection process is divided into two stages: (1) \textbf{Memorization}: adapters are trained to memorize domain-specific KB facts; (2) \textbf{Utilization}: PLMs are trained to leverage adapters when reasoning about entities and their attributes. 

During the memorization stage, adapters are connected to the frozen PLM and tasked with reconstructing corrupted KB facts, thereby memorizing associations between entity and attribute mentions. During the utilization stage, the PLM (now unfrozen) is given access to frozen adapters and learns to leverage their memorized knowledge to make more accurate predictions on downstream tasks such as RG. As a result, PLMs can generalize to unseen inputs by virtue of their domain-general pretraining while receiving domain-specific guidance in their predictions by the knowledge encoded in adapter representations. 

When training KB-adapters, we allocate a single adapter for each individual domain KB (e.g. \textit{hotel} or \textit{restaurant}). This results in shorter training times per adapter and (if needed) facilitates efficient re-training of adapters to reflect changes in the associated KBs.\footnote{E.g. if the user updates the prices of certain items on a restaurant's menu.} This  allows for a straightforward extension of TOD systems equipped with KB-adapters to new domains, as this only requires training a single, new domain-specific adapter that can be used in concert with existing ones. Nevertheless, we also consider a setting where we train a single, mixed-domain adapter on the concatenation of all KBs in our experiments (see \S \ref{multi_setting}).

\subsection{System Overview}
\label{system}

\begin{figure}[t]
\centering
\includegraphics[width=0.9\columnwidth]{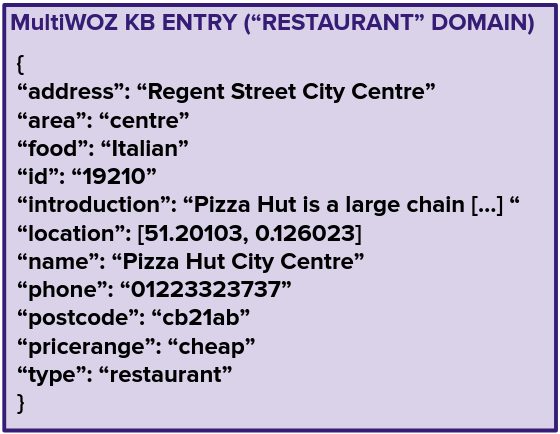}
\caption{Example MultiWOZ 2.2 KB entry.} 
\label{fig:mwoz_kb_example}
\end{figure}

Unlike the vast amounts of data used to pre-train PLMs, information stored in KBs is usually structured and does not resemble natural language expressions. Figure \ref{fig:mwoz_kb_example} shows a single KB entry (or \textit{fact}) from the MultiWOZ 2.2 dataset \cite{budzianowski2018multiwoz,ye2021multiwoz}.
Since KB-adapters need to be compatible with PLMs and their internal representations, we therefore convert KB entries prior to the memorization stage from their initial format into declarative statements of varying complexity (\S \ref{declarative_statements}). Each statement mentions exactly one entity (e.g. a restaurant's name) and one or more entity attributes (e.g. the types of cuisine served by a restaurant). Each statement is subsequently corrupted by masking out a single attribute.\footnote{The entity mention is never masked out, as multiple entities can have the same attribute resulting in ambiguous model inputs, e.g. multiple restaurants can serve \textit{Indian} food.} By denoising the input sequence, adapters learn to correlate entities with their attributes, effectively memorizing entire KBs with high accuracy (\S \ref{memorization_stage}). The obtained KB-adapters are utilized to guide PLMs' predictions during fine-tuning on downstream TOD tasks (\S \ref{exploitation_stage}).

In our experiments, BART \cite{lewis2019bart} is chosen as the PLM that forms the backbone of the adapter-augmented TOD model, due to its competitive performance on generative tasks.\footnote{We utilize the \texttt{BART-Large} provided as part of the \texttt{Transformers} library \cite{wolf2019huggingface}.} While the proposed knowledge injection approach is agnostic to the choice of particular PLM, we leave such validation for future work.

We employ bottleneck adapters \cite{houlsby2019parameter} due to their established effectiveness and insert them after the final layer of the encoder and decoder. The PLM's hidden state given to the adapter as input is combined with the adapter's output using a weighted fusion function which is a linear transformation of the PLM's hidden state followed by a softmax activation that produces the fusion weights. This allows the final model to dynamically adjust the extent to which adapter knowledge is used at each prediction step. In this work, we ran two sets of experiments by applying this gating function to either the logits obtained from both the PLM and the adapters, or to their pre-logit hidden states.

We train a single encoder and a single decoder adapter per domain (hyper-parameter settings are reported in Appendix \ref{sec:hyperparameters}).\footnote{We also investigated several other fusion functions, including unweighted state averaging, state concatenation followed by a projection as used in \cite{wang2020k}, attention, GRU cell, and a combination of softmax distributions produced separately by the PLM and the adapter. However, neither of these performed better than the proposed approach.}

\begin{table}[t]
\smaller[1]
\begin{center}
\begin{tabular}{|p{1.4cm}|p{5.5cm}|}
\hline
~\newline \textbf{atomic facts} & \underline{Pizza Hut City Centre} is located in the \textit{centre} area of the city. \newline \underline{Pizza Hut City Centre} serves food in the \textit{cheap} price range. \newline The postcode of \underline{Pizza Hut City Centre} is \textit{cb21ab}.\newline \ldots \\
\hline
\hline
~\newline \textbf{composite facts} & \underline{Pizza Hut City Centre} is a \textit{restaurant} that serves \textit{Italian} food in the \textit{cheap} price range. It is located at \textit{[51.20103, 0.126023]}, in the \textit{centre} area of the city, in the \textit{cb21ab} postcode. Its phone number is \textit{01223323737}. \\
\hline
\end{tabular}
\caption{Examples of the natural language formats used to represent KB facts in our study. Entity mentions are \underline{underlined}, whereas entity attributes are \textit{italicized}.}
\label{tab: declarative_statements}
\end{center}
\end{table}

\subsection{From KB Facts to Declarative Statements}
\label{declarative_statements}

Previous studies that investigated knowledge injection methods often use relational tuples to represent individual facts contained within a KB, e.g. where an entity is connected to one of its attributes via the relevant relation: \texttt{[Pizza Hut City Centre, food, Italian]}. While this knowledge representation format has been found to be effective in the past, our preliminary studies indicated that the mismatch between the natural language input format expected by a PLM and the structured tuple causes slight performance degradation. Hence, we choose to represent individual KB entries as natural language statements that are fully consistent with the data seen by the PLM during pretraining.  

There are several intuitive ways in which a KB entries can be translated into natural language statements. Referring again to Figure \ref{fig:mwoz_kb_example}, we consider (1) \textbf{atomic} statements, where each statement mentions the entity and one of its attributes, connected by the attribute's relation, and (2) \textbf{composite} statements where each statement communicates the entirety of the entry, covering all provided entity attributes and relations. Table \ref{tab: declarative_statements} illustrates both formats based on the MultiWOZ KB entry in Figure \ref{fig:mwoz_kb_example}. All statements are derived by filling-in pre-defined, human-authored templates with the appropriate entity and attribute values.\footnote{We note that we did not optimize the templates' design as part of our investigation. Our goal in creating the templates was to render structured KB content into natural language without introducing any superfluous information, so as to verify the efficacy of our adapter-based knowledge injection method without additional confounding factors.} Designing the templates introduces minimal overhead, as they reuse attribute designations where possible and do not introduce any information beyond the contents of KB entries. The exhaustive list of templates used in our experiments is provided in Tables \ref{atomic_template_table} and \ref{composite_template_table}. 
During the memorization stage, KB-adapters are trained on a mixture of all atomic and composite facts, so as to familiarize the TOD model with different representations of the same information.

\subsection{Memorization Stage} 
\label{memorization_stage}

\begin{figure}[!t]
\centering
\includegraphics[width=1\columnwidth]{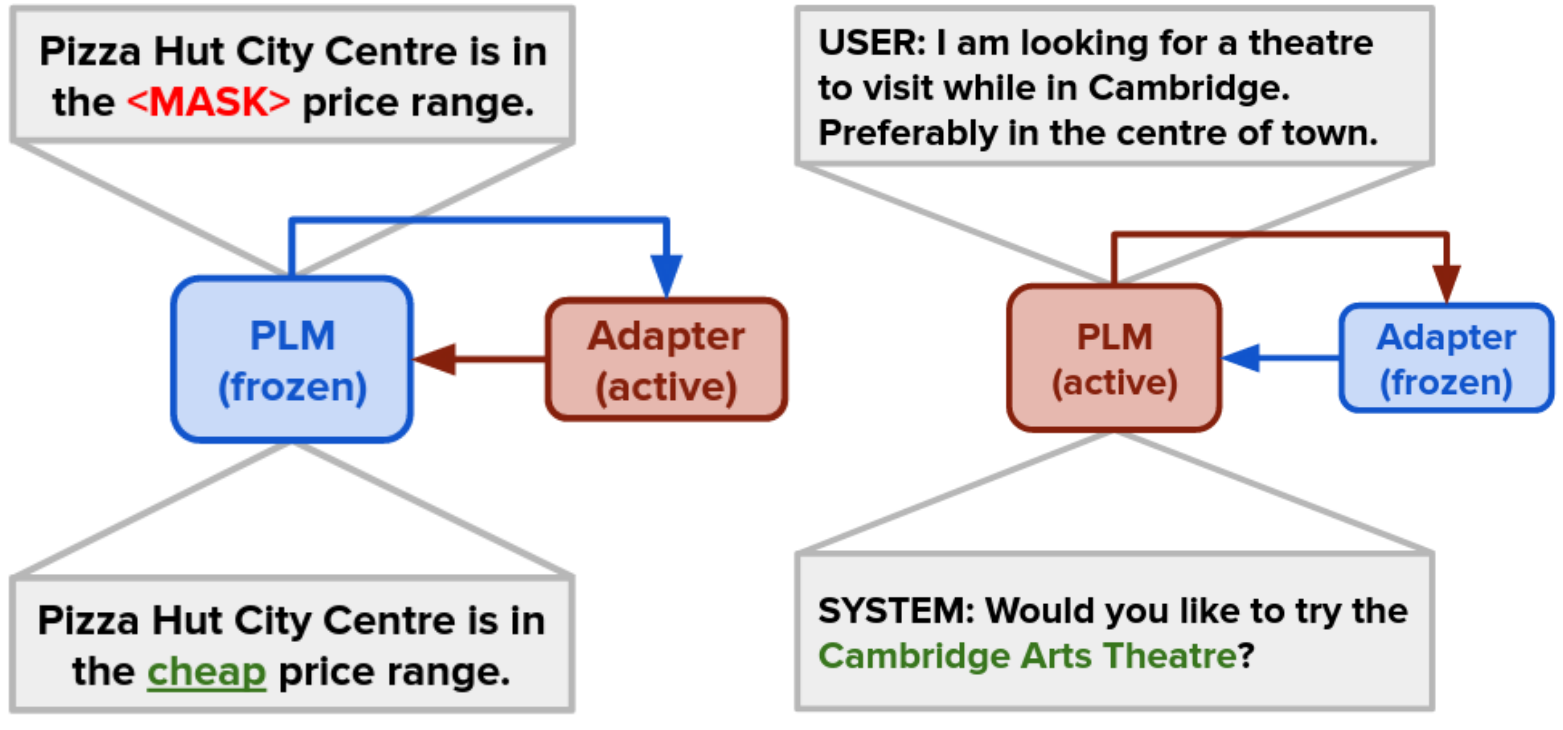}
\caption{On the \textit{left}, a schematic representation of the \textbf{memorization} stage, where the adapter is trained to memorize KB contents by reconstructing corrupted statements derived from KB facts. On the \textit{right}, a representation of the \textbf{utilization} stage, where the adapter-augmented PLM is fine-tuned on a downstream TOD task and learns how to utilize adapter knowledge.}
\label{fig:knowledge_injection_stage}
\end{figure}

Following the construction of natural language representations of KB facts, the memorization stage involves training adapters to memorize and recall KB information. %Its implementation takes inspiration from \cite{wang2020k}. 
As shown on the \textit{left} in Figure \ref{fig:knowledge_injection_stage}, the adapter-augmented PLM learns to reconstruct masked declarative statements that are derived from KB contents, whereby the weights of the PLM itself are kept frozen -- only adapter parameters are being updated. By filling-in masked tokens, adapters learn correlations between entities (e.g. hotel names) and their attributes (e.g. phone numbers). Adapter training resembles masked language modeling and is easy to implement and scale.

\vspace{-1em}
\subsection{Utilization Stage}
\label{exploitation_stage}

After the memorization stage, PLMs are trained to leverage the domain-specific knowledge encoded in adapter representations with the goal of producing more accurate predictions on a downstream task, such as RG, as illustrated on the \textit{right} in Figure~\ref{fig:knowledge_injection_stage}. Throughout this fine-tuning process, adapter parameters are kept frozen so as to preserve the domain-specific knowledge injected during the memorization stage. PLM parameters, on the other hand, are unfrozen to allow the model to learn to exploit adapter representations.

\section{Knowledge-Probing using Response Selection (\KPRS) Benchmark}
\label{knowledge_injection_benchmark}

\begin{figure}[t]
\centering
\includegraphics[width=0.8\columnwidth]{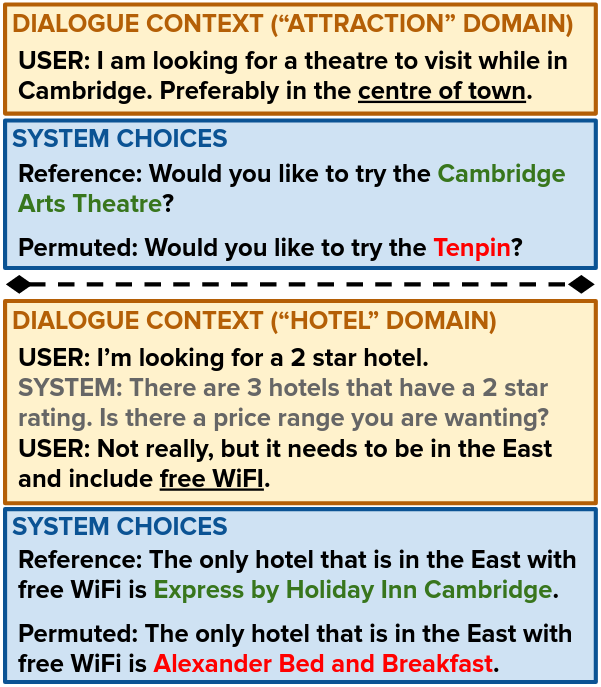}
\caption{Samples from the \KPRS benchmark. Each sample consists of (1) a dialogue context that includes the available history and the active user turn and (2) two candidate responses to be scored by the model -- a reference response that is consistent with both the dialogue context and the KB, and a perturbed response that is not. Reference values are set in \textcolor{newgreen}{\textbf{green}} and perturbed values are set in \textcolor{red}{\textbf{red}}. Note that "Tenpin" is not in the centre area and "Alexander Bed and Breakfast" does not have free WiFi according to their respective KB entries.} 
\label{fig:kprs_examples}
\end{figure}

In this study, we investigate the ability of language models to verify and retrieve domain-specific facts within the TOD setting. To this end, we propose the "\textbf{K}nowledge-\textbf{P}robing using \textbf{R}esponse \textbf{S}election" (\KPRS) task and the associated benchmark. \KPRS allows us to examine whether domain-specific knowledge, such as entities and their attributes, that is stored within the parameters of the evaluated model can be successfully accessed and guide the model's predictions. Being knowledgeable about domain-specific entities in this manner can benefit dialogue models when reasoning about and replying to user queries. We show this to be the case for the response generation task in \S \ref{downstream_results_rg}. 

\KPRS is a contrastive evaluation benchmark that measures whether the probed model has memorized and can accurately retrieve domain-specific knowledge contained within a specified KB. It is derived from MultiWOZ 2.2 dialogues \cite{zang2020multiwoz} (development and test portions only) and covers four domains: \textit{restaurant}, \textit{hotel}, \textit{attraction}, and \textit{train}. Given a dialogue context, the task presented to the evaluated model is to score responses that are either compatible or incompatible with the information contained in the KB.

Importantly, \KPRS should not be regarded as a stand-alone evaluation task, but rather as a probing mechanism that can offer informative insights into a model's ability to access domain-specific facts stored within its parameters, similar to other knowledge probes, e.g. \cite{petroni2019language}. Specifically, a fact-aware model should be able to distinguish between an \textbf{appropriate ("reference") dialogue response} that is compatible with the knowledge base information from an \textbf{inappropriate ("distractor") response} that contradicts the domain-specific knowledge. By design, the two responses are minimally different -- identical except for attribute values associated with entities described in the KB, such as restaurant names or departure times of trains. Hence, to identify the correct dialogue response, a model must be able to distinguish values that are compatible with domain-specific information from those that are not.

\subsection{Benchmark Design}

In order to derive \KPRS from MultiWOZ 2.2 development and test set dialogues, we (1) extract dialogue contexts that precede a system response that contains a mention of an entity from the KB or its attributes, and (2) perturb the corresponding system response to make it incompatible with the KB by modifying said entity and attribute mentions.

Different perturbation strategies are used for different types of attribute slots. For phone numbers, a single digit is randomly changed. For integers (e.g. denoting the price of a train ticket), we randomly increment or decrement the numbers by a small amount. For other slot types, distractor values are chosen so that they differ from the reference value while producing inadmissible responses. Distractors are chosen adversarially, i.e., candidates are sampled from the KB until the perturbed response becomes incompatible with the domain-knowledge and the dialogue context up to the response, while also achieving a lower sentence-level perplexity than the reference response according to a filter-LM (\texttt{BART-Large}). The latter is to ensure the well-formedness and plausibility of the perturbed responses. To guarantee that the perturbed response is indeed unsuitable, we make sure that the selected distractor does not share attriutes that have been mentioned in the dialogue context with the replaced slot value.\footnote{E.g. if the response originally mentioned the name of a restaurant that serves Italian food and the dialogue context up to the response only mentions Italian cuisine as a desired restaurant property, the distractor is explicitly chosen, using string-matching heuristics, to be a restaurant that serves some other type of food, so as not to unintentionally yield a valid response. }  

Figure \ref{fig:kprs_examples} shows examples included in the \KPRS benchmark. Each \KPRS sample contains the \textbf{dialogue context} that includes reference dialogue states, and two response options -- \textbf{reference response} and \textbf{distractor response}. Overall, the \KPRS benchmark dataset includes 3,055 samples (1,711 single-domain, 1,324 multi-domain). Samples had been derived from 831 unique dialogues / 1,997 unique dialogue contexts. On average, 3.65 samples were obtained from each individual dialogue / 1.52 samples from each individual dialogue context. 

\section{Experimental Setup}

\subsection{Knowledge Base Resource}
\label{db_stats}
Throughout our experiments, we use MultiWOZ 2.2 \cite{zang2020multiwoz} which contains several relatively small-scale domain-specific KBs that are aligned with task-oriented dialogues.\footnote{In practical settings, businesses maintain similar knowledge bases in-house which could be utilized in TOD servicees.}
After filtering out KBs with missing information, we are left with four domains: \textit{restaurant}, \textit{hotel}, \textit{attraction}, and \textit{train}. Table \ref{facts} shows the number of facts available per domain. Note the substantial gap in the number of facts where \textit{trains} is approximately 25X to 66X larger than the other domains. 

\begin{table}[!t]
\smaller[1]
\begin{center}
\begin{tabular}{ c c c c}
\hline
restaurant & hotel & attraction & train \\
1,540 &	594	& 1,106	& 39,592 \\
\hline
\end{tabular}
\caption{Number of facts in each KB.}
\label{facts}
\end{center}
\end{table}

\subsection{Intrinsic Evaluation}
\label{fact_memorization_setup}
To examine whether they can accurately retrieve the injected KB facts, we task knowledge-augmented PLMs with reconstructing masked facts, using inputs of the same format as described in \S \ref{declarative_statements}. Since this task measures success as a model's ability to memorize and recall learned KB information rather than generalize it to unseen inputs, we evaluate our models on the same set of data as was used for knowledge injection as part of the memorization stage. Memorization accuracy is employed as the evaluation metric, representing the number of facts that have been correctly reconstructed. We refer to this task as \textit{fact memorization} task.

\subsection{Downstream Evaluation}
Additionally, we evaluate our models on the \KPRS probe (\S \ref{knowledge_injection_benchmark}) as well as the response generation (RG) task. While \KPRS directly estimates  \textbf{models' preference} for dialogue continuations that are either consistent or inconsistent with KB information, RG examines \textbf{model's ability to integrate the injected KB knowledge} into the generated response as part of the TOD pipeline.

For \KPRS, we fine-tune BART-large on the training data for each domain, using correct responses as targets, and evaluate subsequent model performance on the \KPRS benchmark. An augmented PLM that can accurately access the injected domain-specific facts is expected to assign a higher likelihood to the reference response, compared to the permuted distractor. Response selection accuracy is used as the evaluation metric, defined as ($c/N$),
where $N$ is the total number of contrastive sentence pairs and $c$ is the number of pairs in which the reference response (i.e. the one consistent with the KB) is assigned lower perplexity by the model. 

For RG, given a dialogue context, models must generate a response that is consistent with KB facts without performing external KB queries. To test the model's ability for fact retrieval, we use unweighted mean of two informative metrics: inform rate ($n/N$) and success rate ($m/N$) \cite{zang2020multiwoz}, where $N$ is the total number of turns in the test set,  $n$ is the number of turns in which the entities generated by the model are all consistent with the KB, and $m$ is the number of turns in which the model generation provides at least as much of the user-requested information as the gold response.\footnote{BLEU \cite{papineni2002bleu}, as typically used for text generation, is not sufficient as an evaluation metric for our purpose. Previous work evaluated generated responses that contain slot value placeholders instead of concrete information such as entity attributes, as in the case in our study. In addition, any evaluation of response factuality must consider all permissible entities given the dialogue context, rather than only one out of many, as is implicitly done by BLEU.}

\subsection{Baselines}
\label{baseline}
We compare the performance of the knowledge-injected model with two baselines: (1) BART-large without any knowledge augmentation; (2) BART-large that has been sequentially fine-tuned on each KB (Seq-BART). We fine-tune all models on the downstream task prior to the downstream evaluation.

\section{Results \& Analysis}
We examine the models' ability to memorize and retrieve facts learned from the knowledge base in \S \ref{fact_memorization} and the impact of knowledge injection on downstream tasks in \S \ref{downstream_results_KPRS} and \S \ref{downstream_results_rg}. Models were evaluated in the single-domain setting where only one single adapter corresponding to the specified domain was active at evaluation time with test samples belonging exclusively to the adapter domain (a multi-domain setting is discussed in \S \ref{multi_setting})

\subsection{Fact Memorization}
\label{fact_memorization}
As discussed in \S \ref{fact_memorization_setup}, we evaluate whether the knowledge-augmented model is able to successfully denoise masked facts seen during training, thus testing its memorization capabilities. Table \ref{fact_mem} shows the results of the fact memorization task for BART equipped with KB-adapters. The memorization accuracy is generally very high across all domains and appears to correlate with KB size.

\begin{table}[t]
\smaller[1]
\begin{center}
\begin{tabular}{ c c c c c }
\hline
restaurant & hotel & attraction & train \\
98.1 &	98.2 &	97.6 &	93.2 \\
\hline
\end{tabular}
\caption{Fact memorization accuracy for KB-adapters.}
\label{fact_mem}
\end{center}
\end{table}

\subsection{Knowledge-Probing using Response Selection (\KPRS)}
\label{downstream_results_KPRS}
Table \ref{KPRS_results} reports the performance of the knowledge-augmented PLM compared to baselines introduced in \S \ref{baseline}. We found that injecting domain-specific knowledge into the PLM significantly improves \KPRS accuracy -- by 9-15\% -- compared to BART. The largest improvement can be observed in the \textit{train} domain, which is at odds with the fact memorization results (\S \ref{fact_memorization}), where our model underperformed on that domain. As such, while perfect memorization of a of all facts contained within a large KBs remains a challenge in the current training setup, the domain knowledge embedded within the adapter network can nevertheless be effectively exploited by the PLM.

\begin{table}[t]
\smaller[1]
\begin{center}
\begin{tabular}{ l | c c c c | c}
\multicolumn{1}{c|}{Model} & rest. &	hotel &	attr. &	train & all\\
\hline
BART 	&	70.8		&	72.5	&		71.3	&		78.9 & 76.5\\
Seq-BART	&	71.5		&	72.1	&		72.7	&		74.4 & 75.6\\
 \hline
 ada-logits & \textbf{81.5}	& \textbf{83.1}  &	\textbf{81.2}  &	\textbf{94.3} & 78.2 \\
 ada-hidden & 81.3 &	82.0 &	80.6 &	94.0 & \textbf{78.4}\\
\hline
\end{tabular}
\caption{Response selection accuracy on \KPRS . \textit{ada-logits} and \textit{ada-hidden} refer to experiments utilizing KB-adapters with different fusion mechanisms (either at the level of logits or pre-logits hidden states).}
\label{KPRS_results}
\end{center}
\end{table}

\subsection{Response Generation (RG)}
\label{downstream_results_rg}
Presumably, having access to the domain knowledge stored in KB-adapters should enable a PLM to generate responses that are more consistent with the respective KBs. Table \ref{rg_results} reports the results for our RG experiments, providing empirical support for this hypothesis. Interestingly, a large discrepancy can be observed for the \textit{hotel} domain between the two examined representation fusion techniques (\textit{ada-logit} that combines PLM and adapter representations at the logit level vs. \textit{ada-hidden} that combines their pre-logits hidden states). We hypothesise that this is, at least in part, due to the \textit{hotel} KB containing a small number of facts, which may have caused instability during training. Accordingly, although knowledge injection can clearly benefit generation of factual system responses in both the single-domain setting, the extent of the improvements is contingent on the target domain and its properties, as is the best-performing representation combination function.\footnote{It would be valuable to investigate the general impact of KBs' size on the PLMs' performance. However, this falls outside the scope of this paper, as such study would require a greater diversity in the sizes of available KBs.} 

\begin{table}[ht]
\smaller[1]
\begin{center}
\begin{tabular}{ l | c c c c | c}
\multicolumn{1}{c|}{Model} & rest. &	hotel &	attr. &	train & all \\
\hline
% rand-BART	& 0	& 47.0 &	66.1 &	28.2\\
BART 	&	\textbf{54.7}	& 44.3 &	50.3 &	38.2 & 54.2\\
 \hline
 ada-logit &  46.0	& 12.6	& \textbf{69.7} &	\textbf{55.0} & 61.4\\
 ada-hidden & 53.3 &	\textbf{55.9} &	68.6 &	48.6 & \textbf{62.3}\\
\hline
\end{tabular}
\caption{RG performance calculated as the average of inform rate and success rate metrics. The \textit{all} column reports results for the multi-domain setting.}
\label{rg_results}
\end{center}
\end{table}

\label{sec:bleu}

\begin{table}[t]
\smaller[1]
\begin{center}
\begin{tabular}{ l | c c c c | c}
\multicolumn{1}{c|}{Model} & rest. &	hotel &	attr. &	train & all \\
\hline
% rand-BART	& 0	& 47.0 &	66.1 &	28.2\\
BART 	&	12.59	& 11.53 &	15 &	17.71 & 13.41\\
 \hline
 ada-logit & \textbf{12.69}	& 6.5	& 15.09 &	\textbf{19.33} &\textbf{15.98}\\
 ada-hidden & 10.39 &	\textbf{12.64} &	\textbf{15.94} &	17.56 & 15.33\\
\hline
\end{tabular}
\caption{Response generation BLEU score performance.}
\label{rg_results_bleu}
\end{center}
\end{table}

Table \ref{rg_results_bleu} provides estimates of RG quality according to BLEU. Overall, we see minor to substantial improvements with respect to the BLEU metric over the baseline lacking KB-adapters. This can be taken as further evidence in support of the effectiveness of the proposed knowledge injection methodology. However, it should be noted that the extent of the observed improvements varies across domains and representation combination functions.

\subsection{Randomly-initialized Adapters}

We investigate how equipping PLMs with our proposed KB-adapters compares to equipping them with randomly-initialized adapters during the fine-tuning stage (a setting to which we refer as \textit{rand-BART}). This effectively isolates the impact of knowledge injection on the \KPRS and RG performance, by factoring out the increased model capacity due to the additional parameters introduced by the adapters. Table \ref{KPRS_rand_results} shows the experimental results for both tasks. We find that injecting domain-specific knowledge into the PLM does indeed significantly improve \KPRS performance -- by 6-15\% -- compared with \textit{rand-BART}, thus further validating our approach. 

\begin{table}[t]
\smaller[1]
\begin{center}
\begin{tabular}{ l | c c c c}
\multicolumn{1}{c|}{Model} & rest. &	hotel &	attr. &	train \\
\hline
\multicolumn{5}{c}{KPRS} \\
\hline
rand-BART	&	70.3 &	76.6 &	74.4 &	79.6 \\
 \hline
 ada-logits & \textbf{81.5}	& \textbf{83.1}  &	\textbf{81.2}  &	\textbf{94.3} \\
 ada-hidden & 81.3 &	82.0 &	80.6 &	94.0 \\
\hline 
\multicolumn{5}{c}{RG} \\
\hline 
rand-BART	& 0	& 47.0 &	66.1 &	28.2\\
 \hline
 ada-logit &  46.0	& 12.6	& \textbf{69.7} &	\textbf{55.0}\\
 ada-hidden & \textbf{53.3} &	\textbf{55.9} &	68.6 &	48.6\\
\hline
\end{tabular}
\caption{Response selection accuracy on \KPRS and the average of inform and success rate metrics for RG. For RG in the the \textit{restaurant} domain, rand-BART failed to converge given our hyper-parameter settings.}
\label{KPRS_rand_results}
\end{center}
\end{table}

\subsection{Integration of Multiple Knowledge Bases}
\label{multi_setting}

The modular nature of of the proposed knowledge-injection method allows us to equip PLMs with multiple adapters, with each adapter encoding information from a different domain. This enables the augmented PLM to access facts from different domains simultaneously, without running the risk of catastrophic forgetting, whereby information from one domain overwrites previously acquired domain-specific knowledge, e.g. as a result of sequential fine-tuning. Aligned with our motivation to allow users to easily add and modify KBs in practical settings, we investigate whether our proposed system can effectively integrate information from multiple adapters. We utilize the same representation combination functions as described in \S \ref{system}, generalizing them to an unbound number of adapters by computing normalized fusion weights for each adapter and the PLM itself. In this \textit{multi-domain} setting, multiple adapters are active simultaneously, while test samples are drawn from all four studied domains. 

Tables \ref{KPRS_results} and \ref{rg_results} report multi-domain results for \KPRS and RG in the \textit{multi} column. For both tasks, we observe clear improvements compared to baseline models when providing the model with access to all domain-specific adapters simultaneously. However, we note that the gap between the adapter-augmented PLM and the best-performing baseline is much smaller compared to single-domain experiments where the model only has access to a single, relevant adapter (1.9\% vs. 12.25\% on average for \KPRS and 8.1\% vs. 11.6\% on average for RG).

One reason for the limited improvements observed in the multi-domain setting could be the PLM's inability to correctly identify adapters corresponding to the dialogues' domains and to promote their representations. The more pronounced gains observed in the single-domain setting -- where the model does not have to chose between multiple adapters -- appears to support this interpretation. To verify our hypothesis, we preclude the need for adapter selection by instead training a single adapter on the concatenation of facts from all four domains, which preserves the multi-domain setting. Evaluating the performance of the resultant model on \KPRS, we observe improvements over the multiple adapters setting, with \textit{ada-logis} obtaining an accuracy of 83.0\% and \textit{ada-hidden} reaching 85.9\%, thus improving over the best-performing baseline by a substantial \textbf{9.4\%}. This, however, comes at the expense of increased training time during the memorization stage and a significant reduction in flexibility for the addition of new KBs (which will require costly re-training the single, multi-domain adapter rather than simply introducing a new single-domain adapter).

It may be possible to improve the performance of PLMs equipped with multiple single-domain adapters by implementing more expressive combination representation functions or by adjusting the training regime. We regard as a promising research direction that could more effectively extend the flexibility of adapter-based knowledge injection to more complex dialogue settings.

\section{Related Work}

\subsection{Knowledge Injection}
Our work contributes to the growing body of research that explores strategies for introducing external knowledge into the internal reasoning processes of PLMs, with the aim of aligning their predictions with respective knowledge sources \cite{colon2021combining}. Previous work in this area incorporated linguistic \cite{lauscher2019informing, wang2020k}, factual \cite{wang2020k, agarwal2020knowledge}, and commonsense \cite{lauscher2020common} knowledge into pretrained models, with studies differing in the exact format of the injected knowledge and potential modifications to the PLMs' architecture. Nevertheless, injection of highly specific, fine-grained, tabular information commonly associated with TOD (as exemplified by MultiWOZ 2.2 KBs) has so far received limited attention, both within dialogue literature and beyond. The use of natural language statements as the primary mechanism for injecting external information into PLMs has been previously considered in works such as \cite{lu2021kelm}, who trained a generative model to transform knowledge triplets into declarative statements. We rely on template-based generation, instead, to account for the relatively small size of our KBs, the highly structured nature of KB entries, and the lack of natural language sequences that can be trivially aligned with KB contents.

\subsection{Knowledge-Grounded Dialogue}

Of particular relevance to our work is the study by \cite{madotto2020learning} who fine-tune all parameters of a PLM on synthetic dialogues constructed so as to communicate all information contained within a TOD KB. The limitations of their approach, as noted by its authors, are that the synthetic dialogues are noisy and any subsequent updates to the injected KB information require finetuning the entire model anew which is  computationally demanding. We address both issues by relying on grammatically sound templates during knowledge injection and by leveraging light-weight adapters that can be updated for a small fraction of cost incurred by updating the full PLM. The Adapter-Bot introduced in \cite{lin2021adapter} is likewise related to our models in that it employs adapters in the context of TOD. However, rather than training adapters to memorize KB content that can be exploited by the dialogue model without additional supervision, the authors rely on knowledge-aligned dialogues to introduce domain-specific information into their model which may not always be available. More recently, \cite{fan2021augmenting} proposed equipping transformer models with specialized modules that fetch embedded information from external resources, integrating it into the model's reasoning process. While the authors apply their model to dialogue generation, their work differs substantially from ours, as they do not consider the task-oriented setting or structured KBs (instead using training set utterances and Wikipedia excerpts). However, combining knowledge memorization and differential information retrieval is a promising direction for future research. 

Moreover, external knowledge has found application in dialogue literature outside of directly guiding response generation. For instance, \cite{lertvittayakumjorn2021knowledge} annotated dialogue data with constraint violations based on valid links between entities as specified in the corresponding KBs. Similar to \KPRS, detection of constraint violations can be regarded as a probing task that provides insights about the ability of a dialogue model to reason about KB entities.

\section{Limitations}

One of the main limitations of the presented approach is its reliance on manually constructed fact templates. We experimented with fine-tuning KG-adapters directly on $<ent1, rel, ent2>$ KB triples,  but found that the use of templates improves the ability of models to apply the memorized knowledge in downstream applications. In light of this, possible future extensions of our work may include creation of domain-agnostic strategies for knowledge injection that do not necessitate manual design of templates for each new domain. 

Another limitation comes from the fact that the proposed approach is suitable only for static and pseudo-dynamic KBs , i.e. that can change periodically, such as a seasonal menu or a database of cars manufactured by a company. However, it is not suited for real-time databases (e.g. databases that store the availability of rooms in a hotel) since for every KB change the corresponding adapter needs to be retrained in order to be updated. 

Additionally, while injecting knowledge into the language model has been shown to be effective for making it available during fine-tuning on downstream tasks, the knowledge stored in the adapters' parameters might not be accurate enough for certain real world applications due to the imperfect fact memorization we observed in our experiments.
 
Finally, the introduced \KPRS task only evaluates the extent to which a model can access factual information stored in its parameters. It does not not assess the model's ability to understand and use this knowledge for complex reasoning tasks, e.g. counting the number of cars in a specific price range, or listing the items on a menu that do not contain a certain ingredient. This could be an exciting direction for future research.

\vspace{-.2em}
\section{Discussion and Conclusion}
\vspace{-.2em}
In this study, we proposed a method for tightly integrating external knowledge with the internal representations of PLMs by storing domain-specific information within light-weight adapter networks that guide model predictions. Such adapters can memorize KB contents with high accuracy, which decreases slightly for larger KBs. An important contribution of our work is the \KPRS probe designed to measure the ability of TOD models to reason about KB entities and their attributes. As part of our experiments, we showed that KB-adapters clearly benefit the identification and generation of TOD responses that are consistent with dialogue history and relevant KB entries, and showcased the advantages of using adapters for knowledge injection as opposed to sequential fine-tuning. 

Our investigation demonstrates that dialogue models can access domain-specific knowledge without having to query external KBs. This is an important finding as it can pave the way towards reducing the query engineering overhead in chatbot design, thus lowering the entry barrier for developing and deploying real-world TOD systems.

\bibliography{anthology,custom}
\bibliographystyle{acl_natbib}

\clearpage
\appendix

\section{Atomic vs. Compositional Fact Formats}
\label{sec:appendix}

While developing the memorization stage of the knowledge injection process, we compared the relative utility of representing KB facts as either atomic or compositional statements, as measured by the memorization accuracy attained by the adapter-augmented PLM. The results of this pilot experiment are summarized in Table \ref{fact_mem_all}, which paints a mixed picture. While atomic statements result in stronger memorization for the \textit{restaurant}, \textit{hotel}, and \textit{attraction} domains, compositional statements are substantially more effective in the \textit{trains} domain. We therefore decided to combine both formats for our main set of experiments, as the resultant mixture shows reasonable performance across all domains. Furthermore, exposing the model to different surface forms of the same underlying information is expected to enable better generalization for downstream tasks.  

\begin{table}[ht]
\smaller[1]
\begin{center}
\begin{tabular}{ c c c c c c }

 Model & rest &	hotel &	attract &	train \\
  \hline
   atomic & \textbf{98.2} &	\textbf{99.3} &	96.7 &	88.7 \\
composite & 95.8 &	97.6 &	93.7 &	\textbf{97.0} \\
 both & 98.1 &	98.2 &	\textbf{97.6} &	93.2 \\
\hline
\end{tabular}
\caption{Memorization accuracy when training adapters on different formats of declarative statements. \textit{both} denotes the combination of atomic and compositional statements. Scores set in \textbf{bold} are the highest in their respective column.}
\label{fact_mem_all}
\end{center}
\end{table}

\section{Ethical Considerations}
\label{sec:ethic}
Injection of external knowledge into dialogue models may have both ethical and legal implication, if said knowledge contains personal identifiable information (PII), such as social security numbers of addresses of private individuals. Such information would be memorized by the adapter-augmented model and potentially exposed during response generation, if there are no additional safeguards in place to prevent this scenario. For this reason, it is crucial to curate the memorized KBs by removing any and all instances of PII prior to the memorization stage.

\section{Hyper-parameters}
\label{sec:hyperparameters}

All models were trained on V100 GPUs, using the PyTorch implementation of the BART-Large model distributed as part of the HuggingFace Transformers library \cite{wolf2019huggingface}. The training loop employed the AdamW \cite{loshchilov2017decoupled} optimizer. By conducting a grid search, we empirically determined that a learning rate (LR) of $3e^{-5}$ worked best for fine-tuning RG models and LR of $1e^{-6}$ yielded best results for \KPRS. For knowledge injection, LR of $3e^{-5}$ was found to be effective. In all cases, LRs were kept constant across all domains. For all domains and experiments, we re-use the same bottleneck adapter configuration, by setting the size of the hidden layer to $769$. All models were trained until convergence by terminating training after $10$ epochs during which no improvement had been observed on the development set.

\section{Fact Templates}
\label{templates}

This section provides a complete, exhaustive list of all templates used in the generation of declarative statements derived from the MultiWOZ 2.2 KB facts.

\begin{table*}[ht]
\smaller[1]
\begin{center}
\begin{tabular}{ c l l }
Domain & Fact Type & Templates \\
\hline
\multirow{7}{*}{restaurant} & address & The restaurant \{\} is located at \{\}. \\
& area & The restaurant \{\} is located in the \{\} area of the city.' \\
 &  food & The restaurant \{\} serves \{\} food. \\
&  phone & The phone number of the restaurant \{\} is \{\}. \\
&   postcode & The postcode of the restaurant \{\} is \{\}. \\
 &  pricerange & The restaurant \{\} is in the \{\} price range. \\
 &  type & \{\} is a \{\}.\\
 \hline
 \multirow{9}{*}{hotel} & address & The hotel \{\} is located at \{\}.\\
  &      area & The hotel \{\} is located in the \{\} area of the city.\\
  &      internet & The hotel \{\} does\{\}have free internet.\\
  &      parking & The hotel \{\} does\{\}have free parking.\\
  &      phone & The phone number of the hotel \{\} is \{\}.\\
  &      pricerange & The hotel \{\} is in the \{\} price range.\\
  &      stars & The hotel \{\} is rated as \{\} stars.\\
   &     type & The hotel \{\} is a \{\}.\\
   \hline
   
  \multirow{7}{*}{attraction}  & address & The attraction \{\} is located at \{\}.\\
    &  area & The attraction \{\} is located in the \{\} area of the city.\\
    &  entrance fee & The entrance fee for the attraction \{\} is \{\}.\\
    &  phone & The phone number of the attraction \{\} is \{\}.\\
    &  postcode & The postcode of the attraction \{\} is \{\}.\\
    &  pricerange & The attraction \{\} is in the \{\} price range.\\
    &  type & The attraction \{\} is \{\}.\\
\hline
\multirow{7}{*}{train} & arriveBy & The \{\} train arrives at its destination by \{\}.  \\
&        day & The \{\} train operates every \{\}. \\
&        departure & The \{\} train departs from \{\}. \\
&        destination & The destination of the \{\} train is \{\}. \\
&        duration & The duration of the journey with the \{\} train is \{\}. \\
&        leaveAt & The \{\} train leaves at \{\}. \\
&        price & The ticket price for the \{\} train is \{\}. \\
\hline
\end{tabular}
\caption{An exhaustive list of human-authored templates used to generate \textbf{atomic statements} for use in the memorization stage. Note that each domain is allocated exactly one template per entity attribute. Also note that the mask in \textit{does\{\}have} allows for negation in cases where the attribute is negative (e.g. if a hotel does not have free WiFi).}
\label{atomic_template_table}
\end{center}
\end{table*}

\begin{table*}[ht]
\smaller[1]
\begin{center}
\begin{tabular}{ c l }
Domain  & Templates \\
\hline
\multirow{2}{*}{restaurant} & \{\} is a \{\} that serves \{\} food in the \{\} price range. It is located at \\ 
&\{\}, in the \{\} area  of the city, in the \{\} postcode. Its phone number is \{\}.\\
 \hline
 \multirow{2}{*}{hotel} & The hotel \{\} is a \{\} in the \{\} price range. It is rated \{\} stars. It is located at \{\}, in the \{\} area of the city, in the \\
 & \{\} postcode. Its phone number is \{\}. It does\{\}have free parking and it does\{\}have free internet.\\
   \hline
   
\multirow{2}{*}{attraction} & The attraction \{\} is \{\} in the \{\} price range. The entrance fee is \\
& \{\}. It is located at \{\}, in the \{\} area of the city, in the \{\} postcode. Its phone number is \{\}.\\
\hline
\multirow{2}{*}{train} & The \{\} train departs from \{\} every \{\}. It leaves at \{\}. Its destination is \{\} where it arrives at \\
& \{\}.  The duration of the journey is \{\}. The ticket price for this train is \{\}.  \\

\hline
\end{tabular}
\caption{An exhaustive list of human-authored templates used to generate \textbf{composite statements} for use in the memorization stage.}
\label{composite_template_table}
\end{center}
\end{table*}

\end{document}